\documentclass[sigconf]{acmart}
\copyrightyear{2026}
\acmYear{2026}
\setcopyright{cc}
\setcctype{by}
\acmConference[WWW '26]{Proceedings of the ACM Web Conference 2026}{April 13--17, 2026}{Dubai, United Arab Emirates}
\acmBooktitle{Proceedings of the ACM Web Conference 2026 (WWW '26), April 13--17, 2026, Dubai, United Arab Emirates}
\acmPrice{}
\acmDOI{10.1145/3774904.3792653}
\acmISBN{979-8-4007-2307-0/2026/04}

\usepackage{libertine}
\usepackage{booktabs}
\usepackage{multirow}
\usepackage{pifont}
\usepackage{xspace}
\usepackage{enumitem}
\usepackage{balance}
\usepackage[table]{xcolor} 
\usepackage[most]{tcolorbox}
\newcommand{\highlight}{\cellcolor{gray!50}}

\definecolor{customblue}{rgb}{0.168, 0.364, 0.557}
\colorlet{framegray}{gray!3!white}
\newcommand{\dataset}{\textsc{Quit}\xspace}

\begin{document}

\title{Inferential Question Answering}


\author{Jamshid Mozafari}
\authornote{Corresponding Author.}
\orcid{0000-0003-4850-9239}
\affiliation{%
  \institution{University of Innsbruck}
  \city{Innsbruck}
  \country{Austria}
  }
\email{jamshid.mozafari@uibk.ac.at}

\author{Hamed Zamani}
\orcid{0000-0002-0800-3340}
\affiliation{%
  \institution{University of Massachusetts Amherst}
  \city{Amherst}
  \country{United States}
  }
\email{zamani@cs.umass.edu}

\author{Guido Zuccon}
\orcid{0000-0003-0271-5563}
\affiliation{%
  \institution{The University of Queensland}
  \city{Brisbane}
  \country{Australia}
  }
\email{g.zuccon@uq.edu.au}

\author{Adam Jatowt}
\orcid{0000-0001-7235-0665}
\affiliation{%
  \institution{University of Innsbruck}
  \city{Innsbruck}
  \country{Austria}
  }
\email{adam.jatowt@uibk.ac.at}



\begin{abstract}
Despite extensive research on a wide range of question answering (QA) systems, most existing work focuses on answer containment—i.e., assuming that answers can be directly extracted and/or generated from documents in the corpus. However, some questions require \textit{inference}, i.e., deriving answers that are not explicitly stated but can be inferred from the available information. We introduce \textbf{Inferential QA}--a new task that challenges models to infer answers from \emph{answer-supporting} passages which provide only clues. To study this problem, we construct \textbf{\dataset} (\textbf{QU}estions requiring \textbf{I}nference from \textbf{T}exts) dataset, comprising 7,401 questions and 2.4M passages built from high-convergence human- and machine-authored hints, labeled across three relevance levels using LLM-based answerability and human verification. Through comprehensive evaluation of retrievers, rerankers, and LLM-based readers, we show that methods effective on traditional QA tasks struggle in inferential QA: retrievers underperform, rerankers offer limited gains, and fine-tuning provides inconsistent improvements. Even reasoning-oriented LLMs fail to outperform smaller general-purpose models. These findings reveal that current QA pipelines are not yet ready for inference-based reasoning. Inferential QA thus establishes a new class of QA tasks that move towards understanding and reasoning from indirect textual evidence.
\end{abstract}

\ccsdesc[500]{Information systems~Retrieval models and ranking}
\ccsdesc[500]{Information systems~Retrieval tasks and goals}
\ccsdesc[500]{Information systems~Evaluation of retrieval results}


\keywords{Inferential Question Answering, Hints, Retrieval-Augmented Reasoning, Reasoning, Retrieval-Augmented Generation, Large Language Models, Evaluation}


\maketitle

\section{Introduction}\label{s:introduction}

Question Answering (QA) systems~\cite{10.1007/s10115-022-01783-5} aim to provide direct responses to natural language questions. Existing QA tasks include but are not limited to factoid QA \cite{wang2006survey}, descriptive (long-form) and non-factoid QA \cite{lee-etal-2005-descriptive,antique}, Boolean QA \cite{zhang-etal-2024-boolquestions}, and multi-hop QA \cite{yang-etal-2018-hotpotqa}. QA systems have been studied for decades~\cite{pandya2021question} and regardless of which of these tasks they focus on, earlier QA systems primarily focused on \textit{extractive} approaches, which locate an answer span directly in the retrieved passage.
\begin{figure}
  \centering
  \includegraphics[width=0.85\columnwidth]{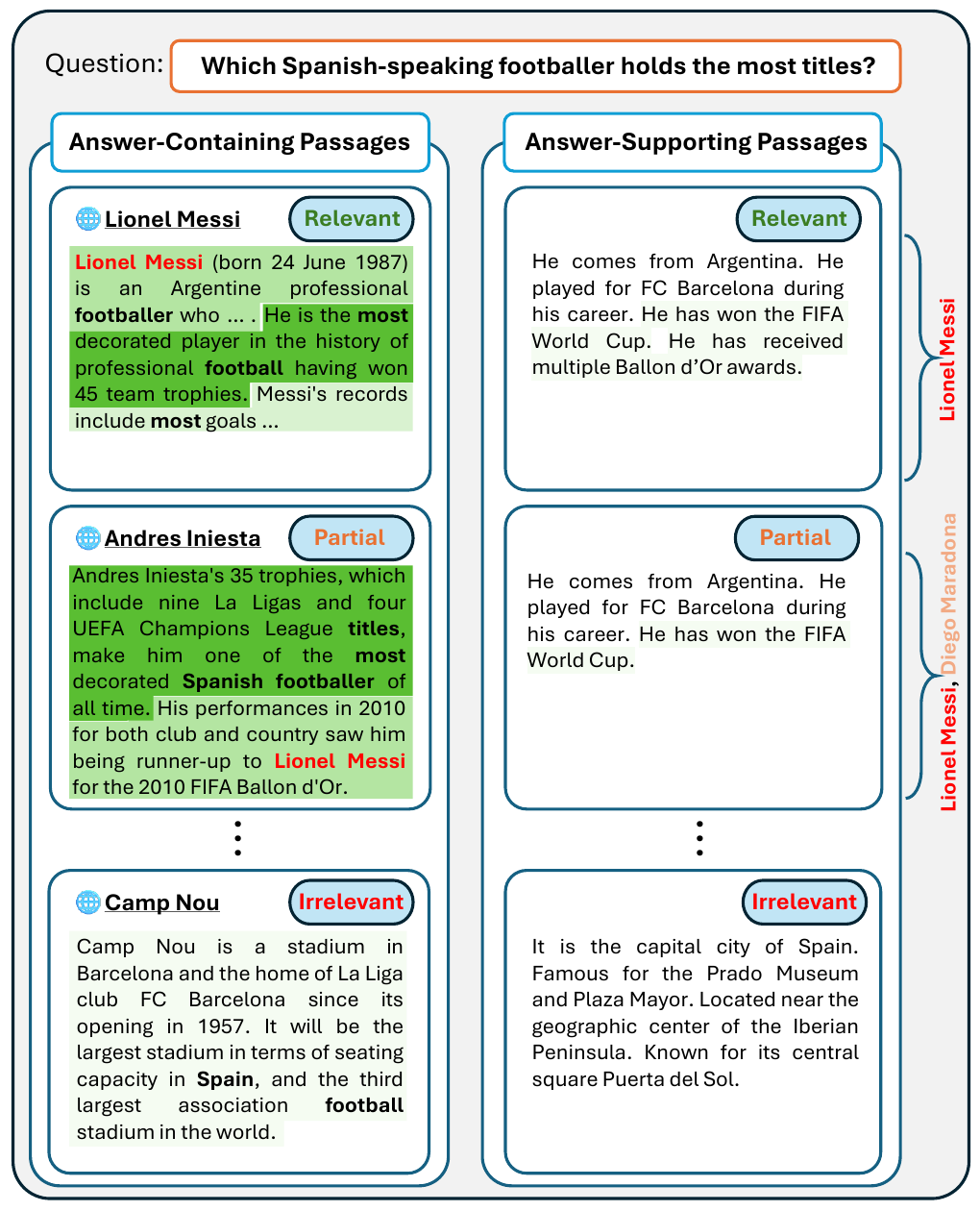}
  \caption{Example of answer-containing passages and answer-supporting passages for the question \textit{Which Spanish-speaking footballer holds the most titles?}, whose answer is \textcolor{red}{Lionel Messi}. \textcolor{green}{Green} highlights indicate semantic similarity between the question and sentences (darker green = higher similarity). \textbf{Bold} words mark lexical overlaps between the question and passages. \textcolor{blue}{Blue rounded rectangles} denote the relevance level of each passage to the question. In the \textit{Answer Supporting} column, the entities that each passage implicitly describes with respect to the question are shown on the right side of the passages.}

  \label{fig:teaser}
\end{figure}
More recent QA systems emphasize on \textit{generative} approaches, where the answer is generated conditioned on the passage content, mostly using large language models (LLMs). Extractive QA systems relied on encoder-based models such as BERT~\cite{devlin-etal-2019-bert} and RoBERTa~\cite{2019arXiv190711692L}.
However, with the advent of decoder-based generative models---including LLMs, such as LLaMA~\cite{2024arXiv240721783D} and Gemma~\cite{gemmateam2025gemma3technicalreport}---the paradigm has shifted toward Retrieval-Augmented Generation (RAG)~\cite{lewis-etal-2020-rag}. This transition is driven by the superior generation and reasoning capabilities of LLMs, which enable them to outperform extractive systems on a broad range of natural language tasks~\cite{minaee2024large}.  

Despite the impressive performance of LLMs on many existing QA benchmarks, we believe that there exists a relatively understudied category of QA tasks that requires further attention in the context of RAG and reasoning---the task of \textbf{Inferential Question Answering (Inferential QA)}. In this task, the answers to the questions do not exist in the corpus; instead, the model must \emph{infer} the correct answer from contextual clues, background knowledge, and logical reasoning supported by the document. For instance, as illustrated in Figure~\ref{fig:teaser}, given the question \emph{``Which Spanish-speaking footballer holds the most titles?''}, the answer-containing relevant passage explicitly states that \emph{Lionel Messi} is the most decorated player, making it directly answerable. In contrast, the answer-supporting relevant passage provides several descriptive clues---that the player is from Argentina, played for FC Barcelona, won the FIFA World Cup, and received multiple Ballon d'Or awards---from which the model must \emph{infer} that the answer is \emph{Lionel Messi}.
This task has real-world applications, such as knowledge-based reasoning systems~\cite{wang-etal-2022-new} and educational tutoring~\cite{barth_et_al_2017}, where deriving answers from indirect or incomplete evidence is essential for evaluating comprehension and reasoning. Furthermore, Inferential QA assesses the reasoning abilities of LLMs and RAG models in settings where answers are not stated explicitly but can be inferred from contextual evidence. Crucially, it differs from commonsense reasoning QA~\cite{talmor-etal-2019-commonsenseqa}, which relies on generic world knowledge. It also differs from multi-hop QA, as it does not require putting multiple passages and entities together to extract or generate the answer.

We first construct \textbf{\dataset} (\textbf{QU}estions requiring \textbf{I}nference from \textbf{T}exts),\footnote{https://github.com/DataScienceUIBK/InferentialQA} a dataset with dedicated training, development, and test splits, containing $7{,}401$ questions and $2{,}405{,}325$ passages.
The passages in \dataset are derived from \textbf{hints}~\cite{10.1145/3578337.3605119, 10.1162/tacl_a_00751}, i.e., clues that guide individuals toward a solution without explicitly providing the answer. By concatenating multiple hints related to an entity, we create passages that describe the entity indirectly. For example, in Figure~\ref{fig:teaser}, the \texttt{relevant}, \texttt{partial}, and \texttt{irrelevant} passages in the Answer-Supported column are formed by concatenating four hints about \textit{Lionel Messi}, three hints about \textit{Lionel Messi} and \textit{Diego Maradona}, and four hints about \textit{Madrid}, respectively. Prior work~\cite{mozafari-etal-2024-exploring} demonstrated that sentence-level hints improve LLM performance in QA compared to retrieved or generated passages, motivating their use in constructing our benchmark. Each passage is annotated with one of three relevance labels: \textbf{2 (fully relevant)} if the passage enables LLM to generate the correct answer, \textbf{1 (partially relevant)} if it references the answer but lacks sufficient information for correct generation, and \textbf{0 (irrelevant)} if it does not support the answer. Annotation details are presented in Section~\ref{sss:passage_labeling}.  

Following the QA literature that uses a Retriever, Reranker, and Reader pipeline for answer extraction or generation, we evaluate diverse retrievers, rerankers, and readers on the Inferential QA task. Our experiments demonstrate that methods effective on existing QA datasets struggle under this setting. In particular, current retrievers and rerankers perform poorly on the Inferential QA task, and fine-tuning yields only limited or inconsistent improvements. These findings highlight that existing QA pipelines are insufficient for inferential questions and retrieval of answer-supporting passages.

To summarize, our main contributions are as follows:
\begin{itemize}[leftmargin=*]
    \item We introduce \textbf{Inferential QA}, a new and challenging QA task that requires models to infer answers from textual clues rather than extract them verbatim, highlighting a gap in current QA paradigms and the limitations of existing pipelines.
    \item We construct \textbf{\dataset} (\textbf{QU}estions requiring \textbf{I}nference from \textbf{T}exts), a large-scale benchmark consisting of $7{,}401$ questions and $2{,}405{,}325$ passages\footnote{Throughout this paper, unless otherwise specified, passage refers to Answer-Supporting passages; Answer-Containing passages are always named explicitly.} derived from hints. The dataset provides dedicated training, development, and test splits, with passages labeled across three levels of relevance.
    \item We comprehensively evaluate retrievers, rerankers, and LLM readers on Inferential QA. Our findings show that (i) methods effective on traditional QA datasets fail to generalize to inference-based reasoning, and (ii) fine-tuning existing models yields only limited or inconsistent improvements.
\end{itemize}


We believe \textbf{Inferential QA} opens a promising direction for retrieval-augmented reasoning models, e.g., Search-R1~\cite{search-r1}, ReasonIR~\cite{shao2025reasonir}, RaDeR~\cite{das-etal-2025-rader}, and DIVER~\cite{long2025diver}, that are capable of \emph{inferring} answers from evidence rather than extracting them, moving QA closer to genuine reasoning and comprehension.


\section{Related Work}\label{s:related_works}

\begin{figure*}[t]
  \centering
  \includegraphics[width=0.85\textwidth]{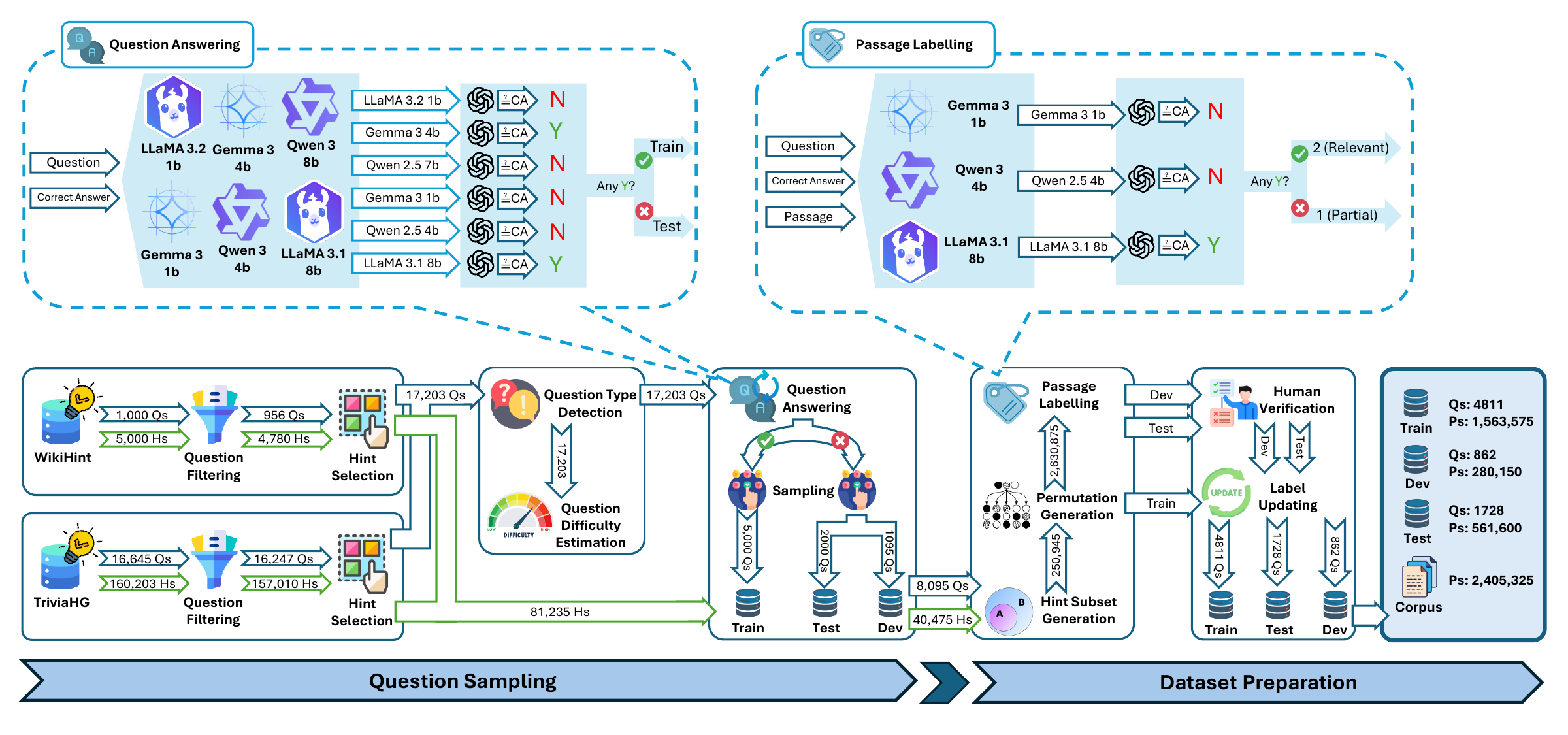}
    \caption{The pipeline for constructing the \dataset dataset. \underline{(1) Question Sampling}: Questions (\textit{Qs}) and their hints (\textit{Hs}) are first filtered to avoid answer leakage. For each question, the top-5 hints are selected based on convergence, followed by question type detection and difficulty estimation. Valid questions are then sampled into temporary training, development, and test sets. \underline{(2) Dataset Preparation}: For each question, subsets and permutations of hints are generated to form diverse passages (\textit{Ps}). These passages are labeled automatically, with dev and test labels further verified and updated through human verification. The final corpus includes 7,401 questions and 2,405,325 passages across training, development, and test splits.}
    \label{fig:pipeline}
\end{figure*}

\subsection{QA Datasets}\label{ss:related_work_qa}
QA research has been driven by large-scale datasets. Early benchmarks such as SQuAD~\cite{rajpurkar-etal-2016-squad}, TriviaQA~\cite{joshi-etal-2017-triviaqa}, WebQuestions~\cite{berant-etal-2013-semantic}, Natural Questions (NQ)~\cite{kwiatkowski-etal-2019-natural}, and BoolQ~\cite{clark-etal-2019-boolq} focus on \textit{answer containment}, where the passage explicitly states the answer. Reasoning-oriented datasets later emerged, including HotpotQA~\cite{yang-etal-2018-hotpotqa} for multi-hop reasoning, DROP~\cite{dua-etal-2019-drop} for numerical reasoning, and StrategyQA~\cite{geva-etal-2021-aristotle} for implicit reasoning. More recent efforts, such as ReasonChainQA~\cite{zhu2022reasonchainqa},
encourage complex inference, while BRIGHT~\cite{su2025bright} emphasizes reasoning-aware retrieval. However, these datasets still assume the presence of passages that explicitly contain the answer. In contrast, \dataset, focuses on answer-supporting passages, where only indirect evidence is available and answers must be inferred.

\subsection{RAG}\label{ss:related_work_rag}
The emergence of LLMs has shifted QA toward generative paradigms, most notably RAG~\cite{lewis-etal-2020-rag}. To improve reasoning in this setting, prior work has explored multi-hop retrieval~\cite{yang-etal-2018-hotpotqa, qi-etal-2021-answering}, iterative retrieval generation cycles~\cite{xu-etal-2023-retrieval}, and prompting strategies that elicit step-by-step reasoning~\cite{wei2022chain}. Despite these advances, RAG methods generally assume that retrieved passages explicitly contain the answer. By contrast, \dataset evaluates QA systems in scenarios where passages provide only indirect evidence, requiring models to infer answers from contextual clues.

\subsection{Inferential Questions}\label{ss:related_work_inferential_qa}
Hints have been recently studied as a way to guide LLMs without directly revealing answers~\cite{mozafari-etal-2024-exploring, mozafari2025hinteval}. Such evidence connects to the notion of \textit{inferential questions}, whose answers cannot be extracted verbatim from given texts but must be inferred from indirect clues, background knowledge, or reasoning. Beyond NLP, inferential questioning has long been emphasized in linguistics and education as central to comprehension and higher-order reasoning~\cite{graesser1994question}. Within NLP, related work has addressed reasoning via multi-hop datasets~\cite{yang-etal-2018-hotpotqa} or implicit commonsense~\cite{geva-etal-2021-aristotle}, yet these still assume that supporting passages explicitly contain the answer. To our knowledge, no dataset benchmarks QA pipelines on passages that only provide indirect evidence. \dataset addresses this gap by constructing large-scale passages from concatenated hints, enabling study of Inferential QA with retrievers, rerankers, and readers.

\section{Methodology}\label{s:methodology}


This section describes the \dataset construction pipeline, including \textit{Question Sampling} and \textit{Dataset Preparation}, as shown in Figure~\ref{fig:pipeline}.

\subsection{Question Sampling}\label{ss:question_sampling}

In this stage, we identify and filter valid questions and their corresponding hints to construct the foundation of the \dataset benchmark.

\subsubsection{Datasets}\label{sss:datasets}

To construct the \dataset benchmark, we use two existing resources that provide question–hint pairs: TriviaHG~\cite{10.1145/3626772.3657855} and WikiHint~\cite{10.1145/3726302.3730284}. TriviaHG contains 16,645 questions and 160,203 automatically generated hints, created using Microsoft Copilot\footnote{\url{https://copilot.microsoft.com/}}, with each question associated with approximately 10 hints. In contrast, WikiHint consists of 1,000 questions and 5,000 human-written hints.
A key feature shared by both datasets is that their hints are designed to describe a concept without explicitly naming it—precisely the type of indirect information needed to construct \textit{answer-supporting passages}. These hints serve as the foundation for generating passages that imply, rather than state, the correct answer.

\subsubsection{Question Filtering}\label{sss:question_filtering}

We begin by filtering out questions for which at least one hint leaks the answer. To identify such cases, we use BEM~\cite{bulian-etal-2022-tomayto} to determine whether any words in the associated hints are semantically equivalent to the correct answer. If at least one word is deemed equivalent, the question is discarded to prevent answer leakage.
Next, for each remaining question, we rank its hints based on their \textit{convergence}~\cite{10.1145/3626772.3657855} scores and retain the top 5 hints\footnote{We select the top 5 hints to maintain computational feasibility, as generating and evaluating all subsets and permutations becomes prohibitively expensive beyond this number.}. Convergence measures how well the hints narrow down the space of possible answers. In other words, it reflects how effectively the hints guide a user toward eliminating incorrect answer candidates and focusing on the correct answer. A high convergence score indicates that a hint strongly points to the entity being the correct answer.
This ensures that we retain only the most informative and accurate hints to construct high-quality passages. After this filtering and selection stage, we obtain a total of 17,203 questions paired with 81,235 high-convergence hints.

\subsubsection{Question Type and Difficulty}\label{sss:question_type_and_difficulty}
To further enrich the dataset with metadata, we detect the type and estimate the difficulty of each question. For question type detection, we use the HintEval framework~\cite{mozafari2025hinteval}, which adopts the classification method proposed by~\citet{tayyar-madabushi-lee-2016-high}. To estimate question difficulty, we apply the Reference-based Question Complexity method~\cite{gabburo-etal-2024-measuring}, which assesses difficulty by analyzing how many of the question’s retrieved passages contain the correct answer and by measuring the semantic relevance between those passages and the question.

\begin{figure}[t]
	\centering
	
	\begin{center}
		\fcolorbox{customblue}{framegray}{
			\begin{minipage}{0.88\linewidth}
				\small
                Question: $<$question$>$ \\
                Answer: $<$ground\_truth$>$ \\
                Candidate: $<$candidate$>$ \\
                Is candidate correct? Choose between "Yes" or "No"
			\end{minipage}
		}
	\end{center}
	    
	\caption{Prompt of GPT-Eval. The placeholder \texttt{<question>} represents the question, \texttt{<ground\_truth>} indicates the correct answer, and \texttt{<candidate>} shows the answer generated by different LLMs.}
	\label{fig:gpt-eval-prompt}
	\Description{}
\end{figure}

\subsubsection{Question Answering}\label{sss:question_answering}

LLMs are capable of answering many questions directly from their parametric knowledge, without relying on external context~\cite{petroni-etal-2019-language}. This poses a challenge for our setup, as it would prevent us from properly evaluating the effect of our passages 
if an LLM can generate the correct answer solely from its parametric knowledge, without relying on any external context. To address this, we filter out questions that the LLMs used as Readers and those used for passage-labeling can correctly answer without any external context.

To reduce bias, we use models from different families and with varying parameter sizes. Specifically, we use Gemma 3 1B~\cite{gemmateam2025gemma3technicalreport}, Qwen 3 4B~\cite{yang2025qwen3}, and LLaMA 3.1 8B~\cite{2024arXiv240721783D} for passage-labeling, and LLaMA 3.2 1B~\cite{2024arXiv240721783D}, Gemma 3 4B~\cite{gemmateam2025gemma3technicalreport}, and Qwen 3 8B~\cite{yang2025qwen3} as Readers\footnote{Other LLMs could also be used; however, we selected these models to create a balanced and fair evaluation environment.}.
Each LLM is prompted with the question alone, without any supporting context. If at least one model produces the correct answer, we classify the question as \textit{parametrically answerable}. Conversely, if none of the models provides the correct answer, the question is classified as \textit{non-parametrically answerable}. 

To compare generated answers with ground-truth answers, we use the GPT-Eval method~\cite{kamalloo-etal-2023-evaluating}, as lexical matching is insufficient in the LLM era. For example, an LLM may output \textit{United States of America} while the gold answer is \textit{USA}; a simple string matching would fail in this case, whereas GPT-Eval correctly recognizes their semantic equivalence. This method uses the prompt shown in Figure~\ref{fig:gpt-eval-prompt}, which outputs \textit{Yes} if the generated answer matches the ground-truth and \textit{No} otherwise. For reliability, we repeat this evaluation three times per question to ensure that none of the selected LLMs can answer it parametrically. 

Finally, we randomly sample 5,000 \textit{parametrically answerable} questions as a temporary training set, since answer leakage\footnote{Answer leakage~\cite{mozafari2025hinteval} measures how much a hint reveals the correct answer inside its content. It ensures that hints guide the user without directly revealing the solution.} is less problematic during training. From the \textit{non-parametrically answerable} pool, we sample 2,000 questions as a temporary test set, with the remaining questions forming the temporary development set. This guarantees that the development and test sets remain high-quality and unbiased for evaluation.

\subsection{Dataset Preparation}\label{ss:dataset_preparation}
In this stage, we generate passages for each question using their hints and label them to make the final \dataset benchmark.

\subsubsection{Subset and Permutation Generation}\label{sss:subset_and_permutation_generation}

From the previous stage, we obtain 8,095 questions and 40,475 hints across the training, development, and test sets. To construct diverse passages, we generate all possible non-empty subsets of the five selected hints for each question. For every subset, we then enumerate all possible permutations of its members. This results in a total of  $\sum_{k=1}^{5} {5 \choose k} k! = 325$ unique passages per question.
For example, the subset containing all five hints yields 120 distinct passages, each expressing the same meaning but differing in sentence order.

\begin{table}
	\footnotesize
	\caption{Statistical Summary of the \dataset Dataset}
	\label{tbl:dataset_statistics}
	\resizebox{\columnwidth}{!}{%
		\scriptsize
		\begin{tabular}{@{}l@{\hspace{90pt}}c@{}}
			\toprule
			Metric                           & Value     \\ \midrule
			Total Questions                  & 7,401     \\
			Total Passages                   & 2,405,325 \\ \midrule
			Avg. Question Length (words)     & 12.56     \\
			Avg. Passage Length (words)      & 58.62     \\
			Avg. Answer Length (words)       & 1.90      \\ \midrule
			Query-Passage Lexical Overlap            & 0.02      \\
			Answer Containment               & 0.00      \\
			Query-Passage Semantic Overlap   & 0.39      \\ \bottomrule
		\end{tabular}%
	}
\end{table}

\subsubsection{Passage Labeling}\label{sss:passage_labeling}

We label all generated passages using LLMs.~\citet{10.1145/3626772.3657957} have demonstrated a strong correlation between passage relevance labels and the answerability of questions by LLMs. Building on this observation, for each question, we provide the question, a candidate passage, and the correct answer.
We pass the question and the passage to three models: Gemma 3 1B, Qwen 3 4B, and LLaMA 3.1 8B to generate an answer based on the passage\footnote{Note that in Section~\ref{sss:question_answering} we evaluated questions without any context, whereas here we include the answer-supporting passages.} using a few-shot prompting strategy\footnote{We use a few-shot strategy to better guide the models toward producing valid answers.} shown in Appendix~\ref{apx:prompt}. To determine whether the answer produced by an LLM is correct, we use GPT-Eval, as described in Section~\ref{sss:question_answering}.
If at least one of the models generates the correct answer from the passage, we label it as \textbf{2 (fully relevant)}, since this indicates that the passage provides sufficient information for an LLM to infer the correct answer. In such cases, we also add the generated answer to the pool of valid answers for that question. Otherwise, we assign the passage the label \textbf{1 (partially relevant)}. The rationale is that even if a passage does not lead directly to the correct answer, it may still provide useful contextual information by indirectly describing the target entity. However, such passages often point to multiple candidate answers rather than uniquely identifying the correct one. For example, in Figure~\ref{fig:teaser} (column \textit{Answer-Supporting}), the \texttt{Relevant} passage enables the generation of \textit{Lionel Messi}, whereas the \texttt{Partial} passage could lead to either \textit{Lionel Messi} or \textit{Diego Maradona}. Additionally, passages containing hints corresponding to
other questions are labeled as \textbf{0 (irrelevant)} for the target question.

\subsubsection{Human Verification}\label{sss:human_verificatin}

To ensure the quality and reliability of the development and test sets, we perform a human verification step. Specifically, we review the answers that GPT-Eval classified as correct, since LLMs—even advanced ones such as GPT—may occasionally produce incorrect or hallucinated outputs. Human validation thus helps us construct more trustworthy development and test sets.  
For this process, we designed an evaluation interface shown in Figure~\ref{fig:human_evaluation} in Appendix~\ref{apx:human_verification} that enables annotators to verify the correctness of answers and resolve potential issues 
effectively.

\subsubsection{Label Updating}\label{sss:label_updating}

During human verification of LLM-generated answers, new cases of answer leakage may inadvertently arise. For example, an answer marked as incorrect by GPT-Eval might later be judged correct by human annotators, and that new correct answer needs to be verified with respect to the risk of answer leakage. To address this issue, we reapply the \textit{Question Filtering} procedure described in Section~\ref{sss:question_filtering} to all questions and remove those for which at least one hint now exhibits answer leakage.  
Human verification may also affect passage labels. Specifically, if a passage no longer yields any correct answers after review, its label is updated from fully relevant (label 2) to partially relevant (label 1). After these adjustments, we obtain the cleaned and finalized versions of the training, development, and test sets, which together constitute the \dataset benchmark. Table~\ref{tbl:dataset_statistics} shows its statistics.

\begin{table}[t]
\footnotesize
\centering
\caption{Retriever performance on \dataset, MS MARCO, and Wikipedia corpora. We report Hit@$k$ ($k \in \{1,10,50,100\}$) and MRR, with the best-performing retriever highlighted.}

\label{tbl:corpus_comparison}
\resizebox{\columnwidth}{!}{%
\scriptsize
\begin{tabular}{@{}l p{50pt} l c c c c c}
\toprule
\textbf{Retriever} & \textbf{Corpus} & \textbf{Hit@1} & \textbf{Hit@10} & \textbf{Hit@50} & \textbf{Hit@100} & \textbf{MRR} \\ \midrule
BM25        & \dataset   & 0.00\%  & 0.25\%  & 0.44\%  & 0.57\%  & 0.04\% \\
            & MS MARCO    & 23.09\% & 41.15\% & 56.77\% & 63.08\% & 29.31\% \\
            & Wikipedia  & 45.60\% & 72.22\% & 84.55\% & 88.25\% & 54.73\% \\ \midrule
DPR         & \dataset   & 9.89\%  & 16.62\% & 21.22\% & 23.74\% & 11.28\% \\
            & MS MARCO    & 14.41\% & 32.47\% & 49.19\% & 58.04\% & 20.55\% \\
            & Wikipedia  & 29.17\% & 57.41\% & 74.65\% & 80.32\% & 38.36\% \\ \midrule
ColBERT     & \dataset   & 12.41\% & 16.44\% & 19.40\% & 20.21\% & 12.62\% \\
            & \highlight MS MARCO    & \highlight 33.28\% & \highlight 49.88\% & \highlight 62.79\% & \highlight 68.63\% & \highlight 38.94\% \\
            & \highlight Wikipedia  & \highlight 60.53\% & \highlight 83.04\% & \highlight 90.86\% & \highlight 92.94\% & \highlight 68.86\% \\ \midrule
Contriever  & \dataset   & 6.49\%  & 13.29\% & 18.95\% & 22.54\% & 8.15\% \\
            & MS MARCO    & 19.85\% & 43.63\% & 60.30\% & 66.72\% & 28.06\% \\
            & Wikipedia  & 45.37\% & 76.79\% & 87.44\% & 90.80\% & 56.56\% \\ \midrule
BGE         & \highlight \dataset   & \highlight 12.85\% & \highlight 21.98\% & \highlight 27.96\% & \highlight 30.23\% & \highlight 14.68\% \\
            & MS MARCO    & 27.55\% & 47.69\% & 63.31\% & 71.93\% & 34.35\% \\
            & Wikipedia  & 53.47\% & 81.71\% & 91.72\% & 94.33\% & 63.31\% \\ \bottomrule
\end{tabular}%
}
\end{table}

\section{Experimental Setup}\label{s:experimental_setup}

In our experiments, we evaluate diverse retrievers, rerankers, and readers on the Inferential QA task and \dataset benchmark.

For the \textit{retriever}, we employ five retrieval methods: BM25~\cite{10.1561/1500000019}, DPR~\cite{karpukhin-etal-2020-dense}, ColBERT v2~\cite{santhanam-etal-2022-colbertv2}, Contriever~\cite{izacard2022unsupervised}, and BGE~\cite{chen2024bge}, all accessed via the Pyserini toolkit~\cite{10.1145/3404835.3463238}. We fine-tune DPR using the Tevatron library~\cite{10.1145/3539618.3591805}, and ColBERT using its official repository\footnote{\url{https://github.com/stanford-futuredata/ColBERT}}.

For the \textit{reranker}, we experiment with five models: LiT5 (T5-3B)~\cite{tamber2023scaling}, MonoT5 (T5-3B)~\cite{nogueira-etal-2020-document}, RankGPT (LLaMA 3.1 8B)~\cite{sun-etal-2023-chatgpt}, RankT5 (T5-3B)~\cite{10.1145/3539618.3592047}, and UPR (T0-3B)~\cite{sachan-etal-2022-improving}, implemented using the Rankify~\cite{abdallah2025rankify}. MonoT5 is fine-tuned using the PyGaggle~\cite{10.1007/978-3-031-28241-6_10}.

For the \textit{reader}, we employ three LLMs from different model families and parameter scales: LLaMA 3.2 1B~\cite{2024arXiv240721783D} and Gemma 3 4B~\cite{gemmateam2025gemma3technicalreport} as general-purpose LLMs, and Qwen 3 8B~\cite{yang2025qwen3} as a reasoning-oriented model. This setup allows us to evaluate the effectiveness of LLMs with different specializations on inferential questions. As noted in Section~\ref{sss:question_answering}, we choose these models because we verified that none of them can answer the questions solely from its parametric knowledge, which ensures a fair evaluation on passages. All models are executed on two NVIDIA A40 GPUs with 40GB.

\begin{table}[t]
\centering
\caption{Comparison of rerankers across \dataset, MS MARCO, and Wikipedia corpora. We report Hit@$k$ ($k=1,10,50$) and MRR. The best reranker for each corpus is highlighted, while the highest score is \textbf{bolded}.}
\label{tbl:corpus_reranker}
\resizebox{\columnwidth}{!}{%
\scriptsize
\begin{tabular}{@{}l l l c c c c}
\toprule
\textbf{Retriever} & \textbf{Corpus} & \textbf{Reranker} & \textbf{Hit@1} & \textbf{Hit@10} & \textbf{Hit@50} & \textbf{MRR} \\ \midrule
BGE       & \dataset   & LiT5        & 26.03\% & 29.21\% & \textbf{33.80\%} & 28.10\% \\
          &            & \highlight MonoT5      & \highlight \textbf{27.60\%} & \highlight 29.98\% & \highlight 32.35\% & \highlight \textbf{28.54\%} \\
          &            & RankGPT     & 24.02\% & 29.05\% & 33.33\% & 25.70\% \\
          &            & RankT5      & 26.62\% & \textbf{30.44}\% & 32.52\% & 27.80\% \\
          &            & UPR         & 26.85\% & 29.86\% & 32.70\% & 27.89\% \\ \midrule
ColBERT   & MS MARCO    & LiT5        & \textbf{40.28\%} & 53.53\% & 64.24\% & 44.80\% \\
          &            & MonoT5      & 38.60\% & 55.44\% & 65.10\% & 44.27\% \\
          &            & RankGPT     & 28.88\% & 49.88\% & 62.79\% & 36.76\% \\
          &            & \highlight RankT5      & \highlight 39.53\% & \highlight \textbf{55.56\%} & \highlight \textbf{65.39\%} & \highlight \textbf{45.01\%} \\
          &            & UPR         & 36.40\% & 54.69\% & 65.22\% & 42.62\% \\ \midrule
ColBERT   & Wikipedia  & LiT5        & 68.52\% & 85.82\% & 91.38\% & 74.59\% \\
          &            & MonoT5      & 68.17\% & 88.02\% & 91.84\% & 75.33\% \\
          &            & RankGPT     & 52.60\% & 82.99\% & 90.86\% & 64.60\% \\
          &            & \highlight RankT5      & \highlight \textbf{71.88\%} & \highlight \textbf{88.48\%} & \highlight 91.96\% & \highlight \textbf{78.06\%} \\
          &            & UPR         & 65.57\% & 86.52\% & \textbf{92.13\%} & 73.37\% \\ \bottomrule
\end{tabular}%
}
\end{table}

\begin{figure}[t]
  \centering
  \includegraphics[width=0.75\columnwidth]{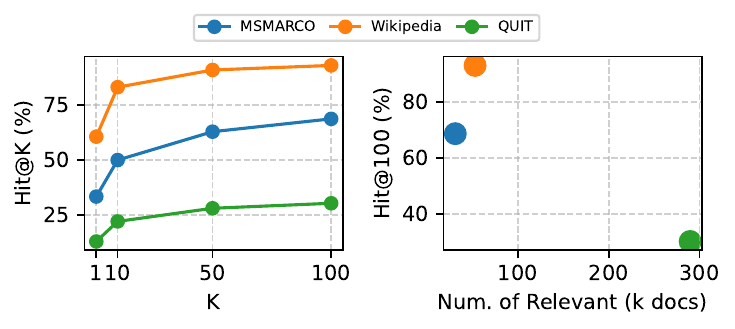}
    \caption{Left: Hit@$k$ comparison across different corpora. Right: Hit@100 with respect to the number of relevant passages (label~2) across corpora.}
    \label{fig:corpuses_comparison}
\end{figure}

We evaluate our experiments using standard metrics: \textbf{Hit@$k$}, which measures whether a gold passage appears in the top-$k$ (reported for $k=1, 5, 10, 50, 100$);
\textbf{Recall@$k$}, which evaluates the coverage of relevant passages (for $k=5, 10, 50$); \textbf{MRR}, the mean reciprocal rank of the first relevant passage; \textbf{NDCG@$k$}, which captures ranking quality with graded relevance (for $k=10$ and $k=100$); and \textbf{EM}, the exact match between predicted and gold answers in the reader component.

\section{Results and Discussion} \label{s:results_and_discussion}

\begin{table}[t]
\small
\centering
\caption{Retriever performance on vanilla and finetuned models. We report Hit@$k$, Recall@$k$, MRR, and nDCG@$k$.}
\label{tbl:retriever_results}
\resizebox{\columnwidth}{!}{%
\scriptsize
\begin{tabular}{@{}l c c c c c c c c}
\toprule
\textbf{Retriever}
& \multicolumn{2}{c}{Hit@K}
& \multicolumn{3}{c}{Recall@K}
& MRR@K
& \multicolumn{2}{c}{nDCG@K} \\
\cmidrule(lr){2-3} \cmidrule(lr){4-6} \cmidrule(lr){7-7} \cmidrule(lr){8-9}
& 1 & 5
& 5 & 10 & 50
& 100
& 10 & 100 \\ \midrule

\multicolumn{9}{l}{\textbf{Vanilla}} \\
BGE        &  23.73\% &  27.37\% &  0.37\% &  0.75\% &  3.80\% &  25.45\% &  18.95\% &  21.14\% \\
DPR        & 18.58\% & 21.18\% & 0.28\% & 0.54\% & 2.54\% & 19.77\% & 13.68\% & 13.64\% \\
ColBERT    &  20.31\% &  21.12\% &  0.31\% &  0.63\% &  3.15\% &  20.69\% &  16.14\% &  17.51\% \\ \midrule

\multicolumn{9}{l}{\textbf{Finetuned}} \\
FT-DPR     &  20.91\% &  28.07\% &  0.32\% &  0.63\% &  3.12\% &  23.56\% &  14.98\% &  16.69\% \\
FT-ColBERT & 21.11\% &  23.84\% & 0.27\% & 0.53\% & 2.56\% &  20.83\% & 13.00\% & 13.98\% \\ \bottomrule
\end{tabular}%
}
\end{table}

\subsection{Compared Benchmarks}\label{baselines}
To evaluate how challenging the Inferential QA task is for current retrievers and rerankers, we compare \dataset corpus\footnote{We refer to the union of all of answer-supporting passages from the training, development, and test sets as the \dataset corpus.} against two widely used benchmarks: Wikipedia~\cite{karpukhin-etal-2020-dense} and MS MARCO~\cite{bajaj2016ms}. Specifically, we retrieve passages from all three corpora (\dataset, Wikipedia, and MS MARCO) for the questions of the test set. 

In traditional QA systems, relevance is typically defined by answer containment, i.e., a passage is considered relevant if it includes the correct answer. The \textit{Answer-Containing} column in Figure~\ref{fig:teaser} illustrates this definition. However, as explained in Section~\ref{sss:passage_labeling}, our labeling method differs because our passages are designed to avoid answer leakage, making answer containment unsuitable. To address this, we relabel the retrieved passages from Wikipedia and MS MARCO using our method. 
Since partially relevant (label 1) in our framework is defined based on the relationship between a hint and a question, it cannot be directly applied to these corpora, which lack hints, we therefore use passages labeled as fully relevant (label 2), which are determined based on whether an LLM can correctly answer the question using the passage, 
to ensure a fair comparison with answer-supporting passages. 

For fairness, we retrieve the top 100 passages from \dataset, MS MARCO, and Wikipedia using five different retrievers, and the results are reported in Table~\ref{tbl:corpus_comparison}. These results reveal a large gap in Hit@$k$ and MRR between \dataset and the other corpora, highlighting the inherent difficulty of retrieving relevant passages in the Inferential QA task. The table also shows that the best-performing retriever for \dataset differs from those for MS MARCO and Wikipedia, further underscoring the difference between answer-supporting and answer-containing passages. Figure~\ref{fig:corpuses_comparison} further illustrates this gap. The left chart shows that Hit@$k$ is consistently highest for Wikipedia, followed by MS MARCO, and lowest for \dataset, demonstrating the increased challenge of retrieving from \dataset. The right chart shows that although the retrieved passages of \dataset include 289{,}660 relevant passages (label~2), Hit@100 is only 30.23\%. In contrast, for MS MARCO and Wikipedia, which contain 31,195 and 52,945 relevant retrieved passages, Hit@100 reaches 68.63\% and 92.94\%, respectively. This indicates that, although one might expect a higher number of relevant passages to yield a higher Hit@100, this is not the case for \dataset. Finally, it is worth noting that the average passage length in MS MARCO is 56.57 tokens, while in \dataset it is 58.62 tokens, confirming that the observed gap is not due to differences in passage length.

In addition to retrievers, Table~\ref{tbl:corpus_reranker} shows that the best reranker for \dataset differs from those for MS MARCO and Wikipedia, indicating that answer-supporting passages behave differently from answer-containing passages and that the prevailing hypothesis about what constitutes the most relevant passages does not hold for answer-supporting passages.

\begin{table}[t]
\small
\centering
\caption{Performance comparison of rerankers applied to BGE and finetuned retrievers. We report Hit@K, Recall@K, MRR, and nDCG@K. The best reranker for each retriever is highlighted, and the highest score in each column is \textbf{bolded}.}
\label{tbl:rerankers_best_finetuned}
\resizebox{\columnwidth}{!}{%
\scriptsize
\begin{tabular}{@{}l l c c c c c c c c}
\toprule
\textbf{Retriever} & \textbf{Reranker}
& \multicolumn{2}{c}{Hit@K}
& \multicolumn{3}{c}{Recall@K}
& \multicolumn{1}{c}{MRR@K}
& \multicolumn{2}{c}{nDCG@K} \\
\cmidrule(lr){3-4} \cmidrule(lr){5-7} \cmidrule(lr){8-8} \cmidrule(lr){9-10}
&
& 1 & 5
& 5 & 10 & 50
& 100
& 10 & 100 \\ \midrule

\multirow{5}{*}{BGE}
 & LiT5     & 27.03\% & 29.17\% & 0.41\% & 0.82\% & 3.86\% & 28.10\% & 21.15\% & 21.49\% \\
 & MonoT5   & \highlight \textbf{27.60\%} & \highlight \textbf{29.46\%} & \highlight \textbf{0.42}\% & \highlight \textbf{0.84\%} & \highlight \textbf{4.01\%} & \highlight \textbf{28.54\%} & \highlight \textbf{22.36\%} & \highlight \textbf{21.81\%} \\
 & RankGPT  & 24.02\% & 27.49\% & 0.37\% & 0.75\% & 3.78\% & 25.70\% & 18.98\% & 21.14\% \\
 & RankT5   & 26.62\% & 29.05\% & 0.41\% & 0.82\% & 4.00\% & 27.80\% & 21.67\% & 21.69\% \\
 & UPR      & 26.85\% & 28.94\% & 0.42\% & 0.83\% & 4.01\% & 27.89\% & 21.06\% & 21.61\% \\ \midrule

\multirow{5}{*}{FT\text{-}DPR}
 & LiT5     & 27.84\% & 32.58\% & 0.40\% & 0.77\% & 3.08\% & 30.34\% & 19.31\% & 16.20\% \\
 & MonoT5   & \highlight \textbf{28.01\%} & \highlight 31.89\% & \highlight 0.40\% & \highlight \textbf{0.78}\% & \highlight \textbf{3.39}\% & \highlight 30.24\% & \highlight \textbf{20.34\%} & \highlight \textbf{16.63}\% \\
 & RankGPT  & 20.25\% & 28.41\% & 0.31\% & 0.61\% & 2.90\% & 24.55\% & 14.21\% & 15.36\% \\
 & RankT5   & 26.97\% & 31.71\% & 0.40\% & 0.78\% & 3.39\% & 29.60\% & 19.98\% & 16.56\% \\
 & UPR      & 27.89\% & \textbf{32.64\%} & \textbf{0.41\%} & 0.78\% & 3.37\% & \textbf{30.38\%} & 19.60\% & 16.49\% \\ \midrule

\multirow{5}{*}{FT\text{-}ColBERT}
 & LiT5     & \textbf{23.32\%} & 25.41\% & 0.33\% & 0.63\% & 2.66\% & 24.53\% & 16.37\% & 14.52\% \\
 & MonoT5   & \highlight 22.69\% & \highlight 25.58\% & \highlight \textbf{0.33}\% & \highlight \textbf{0.65\%} & \highlight \textbf{2.86\%} & \highlight 24.16\% & \highlight \textbf{17.07\%} & \highlight \textbf{14.80\%} \\
 & RankGPT  & 18.00\% & 23.84\% & 0.27\% & 0.53\% & 2.55\% & 20.75\% & 13.05\% & 13.98\% \\
 & RankT5   & 22.92\% & 25.52\% & 0.33\% & 0.64\% & 2.84\% & \textbf{24.29\%} & 16.82\% & 14.74\% \\
 & UPR      & 22.80\% & \textbf{25.81\%} & 0.33\% & 0.63\% & 2.84\% & 24.33\% & 16.31\% & 14.67\% \\ 
\bottomrule
\end{tabular}%
}
\end{table}

\begin{table}[t]
\small
\centering
\caption{Performance of retrievers with the finetuned reranker (FT-MonoT5). We report Hit@$k$, Recall@$k$, MRR, and nDCG@$k$ for both vanilla and finetuned retrievers.}
\label{tbl:retrievers_finetuned_reranker}
\resizebox{\columnwidth}{!}{%
\scriptsize
\begin{tabular}{@{}l c c c c c c c c@{}}
\toprule
\textbf{Retriever}
& \multicolumn{2}{c}{Hit@K}
& \multicolumn{3}{c}{Recall@K}
& MRR@K
& \multicolumn{2}{c}{nDCG@K} \\
\cmidrule(lr){2-3} \cmidrule(lr){4-6} \cmidrule(lr){7-7} \cmidrule(lr){8-9}
& 1 & 5
& 5 & 10 & 50
& 100
& 10 & 100 \\ \midrule

\multicolumn{9}{l}{\textbf{Vanilla}} \\
BM25       & 0.00\% & 0.06\% & 0.00\% & 0.00\% & 0.00\% & 0.05\% & 0.02\% & 0.01\% \\
Contriever & 12.69\% & 16.53\% & 0.18\% & 0.34\% & 1.53\% & 14.69\% & 8.95\% & 8.27\% \\
BGE        & 23.44\% & 26.98\% & 0.37\% & 0.74\% & 3.76\% & 25.10\% & 18.67\% & 20.77\% \\
DPR        & 18.29\% & 20.88\% & 0.28\% & 0.53\% & 2.50\% & 19.47\% & 13.50\% & 13.37\% \\
ColBERT    & 20.07\% & 20.84\% & 0.31\% & 0.62\% & 3.11\% & 20.43\% & 15.89\% & 17.21\% \\ \midrule

\multicolumn{9}{l}{\textbf{Finetuned}} \\
FT-DPR     & 19.91\% & 28.07\% & 0.30\% & 0.60\% & 2.93\% & 24.07\% & 13.96\% & 15.36\% \\
FT-ColBERT & 18.11\% & 23.84\% & 0.27\% & 0.53\% & 2.56\% & 20.83\% & 13.00\% & 13.98\% \\ \bottomrule
\end{tabular}%
}
\end{table}

\subsection{Retriever}\label{ss:retriever}
To evaluate the performance of retrievers for the Inferential QA task with respect to both partially relevant (label~1) and fully relevant (label~2) passages, we test BGE (the best-performing retriever for \dataset, as shown in Table~\ref{tbl:corpus_comparison}) alongside DPR and ColBERT, each in both vanilla and fine-tuned variants. This setup allows us to analyze the impact of fine-tuning on retrieving answer-supporting passages.

For fine-tuning DPR and ColBERT, we experiment with various numbers of positive and negative training samples ($1$, $5$, $10$, $50$, $100$, and $200$), inspired by the findings of~\citet{chang2025improving}, who showed that increasing the number of positive samples during training can enhance retrieval performance. We find that the optimal configuration is 10 positives (and 10 negatives\footnote{We use the same number of positive and negative passages, and negatives are sampled from the positive passages of other questions.}) for DPR, and 50 positives for ColBERT. Full results across all configurations are reported in Table~\ref{tbl:finetuned_colbert} and Table~\ref{tbl:finetuned_dpr} in Appendix~\ref{apx:finetuned_retrievers}.

As shown in Table~\ref{tbl:retriever_results}, BGE consistently achieves the highest performance. Fine-tuning leads to a slight improvement for DPR but provides no noticeable gain for ColBERT. We hypothesize that this difference stems from architectural design: DPR employs two separate BERT encoders for questions and passages, which may help it better capture passages. In contrast, ColBERT relies on a single shared encoder with late interaction, which may limit its ability to adapt to the challenges of the Inferential QA task. Also, BGE’s strong performance may be attributed to its LLM-based architecture, which is better equipped to model deeper semantic connections and implicit clues—key characteristics of answer-supporting 
passages. 

Overall, these results suggest that \textit{fine-tuning offers only marginal benefits and is insufficient for tackling the complexity of the Inferential QA task. Addressing this challenge may require fundamentally new retrieval approaches and paradigms.}

\begin{table}[t]
\small
\centering
\caption{Performance of rerankers under the oracle setting. We report nDCG@$k$ ($k=5,10,50,100$) for both vanilla and finetuned rerankers. The best results are highlighted.}
\label{tbl:oracle_results}
\resizebox{\columnwidth}{!}{%
\scriptsize
\begin{tabular}{@{} p{100pt} c c c c}
\toprule
\textbf{Reranker}
& \multicolumn{4}{c}{nDCG@K} \\
\cmidrule(lr){2-5}
& 5 & 10 & 50 & 100 \\ \midrule

\multicolumn{5}{l}{\textbf{Vanilla}} \\
LiT5     & 72.94\% & 75.49\% & 79.34\% & 82.99\% \\
RankGPT  & 65.02\% & 69.74\% & 78.09\% & 82.24\% \\
RankT5   & 78.96\% & 80.18\% & 84.69\% & 87.49\% \\
UPR      & 78.56\% & 79.72\% & 84.30\% & 87.25\% \\
MonoT5   & 82.01\% & 82.95\% & 86.46\% & 88.71\% \\ \midrule

\multicolumn{5}{l}{\textbf{Finetuned}} \\
FT-MonoT5 & \highlight 83.56\% & \highlight 84.24\% & \highlight 87.08\% & \highlight 89.17\% \\ 
\bottomrule
\end{tabular}%
}
\end{table}

\begin{figure}[t]
  \centering
  \includegraphics[width=0.5\columnwidth]{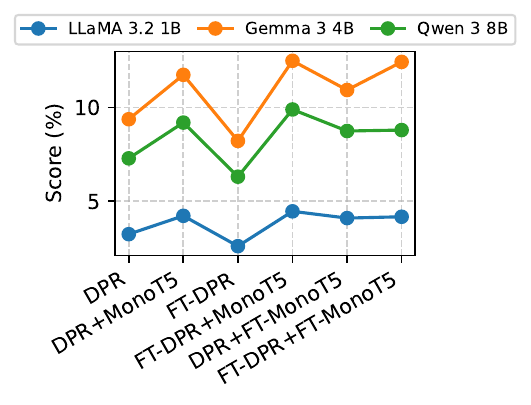}
    \caption{Trend of EM across different settings using DPR (vanilla and fine-tuned) as the retriever and MonoT5 (vanilla and fine-tuned) as the reranker, evaluated with LLaMA 3.2 1B, Gemma 3 4B, and Qwen 3 8B as readers in the RAG setup.}
    \label{fig:rag_results}
\end{figure}

\subsection{Reranker}\label{ss:reranker}

To evaluate the effect of rerankers on the Inferential QA task, we test the \dataset benchmark using five different rerankers. These rerankers are applied on top of BGE (the best retriever) as well as the fine-tuned versions of DPR and ColBERT, allowing us to analyze their impact on this task. Table~\ref{tbl:rerankers_best_finetuned} shows that MonoT5 achieves the best performance with BGE and is also the top reranker for FT-DPR and FT-ColBERT. Compared to the retriever results in Table~\ref{tbl:retriever_results}, rerankers provide only marginal improvements, indicating that reranking alone is not sufficient to reliably promote passages to the top of the retrieved list. \textit{This suggests that understanding answer-supporting passages is also challenging for rerankers.}

In addition to the vanilla rerankers, we fine-tune MonoT5 under different settings, varying the number of positive and negative samples as in Section~\ref{ss:retriever}. We find that the best configuration uses 10 positive passages, with detailed results provided in Table~\ref{tbl:finetuned_monot5} in Appendix~\ref{apx:finetuned_reranker}.
However, as shown in Table~\ref{tbl:retrievers_finetuned_reranker}, the fine-tuned MonoT5 performs worse than its vanilla counterpart (Table~\ref{tbl:rerankers_best_finetuned}), indicating that fine-tuning does not help rerankers adapt to the Inferential QA task.
Overall, \textit{similar to retrievers, rerankers require new approaches and paradigms to effectively handle answer-supporting passages.}

\begin{table}[t]
\small
\centering
\caption{LLM-based Reader results across different retriever--reranker strategies. We report Exact Match (EM) for three LLMs. \textit{UN} refers to $\text{Union}_{norm}$ and \textit{UF} to $\text{Union}_{freq}$ methods.}
\label{tbl:rag_results}
\resizebox{\columnwidth}{!}{%
\scriptsize
\begin{tabular}{@{}l l c c c c c c@{}}
\toprule
\textbf{Retriever} & \textbf{Reranker} & \textbf{K} & \textbf{Strategy}
& \textbf{LLaMA-32-1b} & \textbf{Gemma-3-4b} & \textbf{Qwen-3-8b} \\
\midrule

\multicolumn{7}{l}{\textbf{Optimal}} \\
Oracle & Oracle & Oracle & Oracle
& 40.68\% & 90.16\% & 62.50\% \\ \midrule

\multicolumn{7}{l}{\textbf{Oracle Retriever + Reranker (Vanilla \& Finetuned)}} \\
Oracle & MonoT5     & 5 & UF & 20.25\% & 50.41\% & 34.32\% \\
Oracle & FT-MonoT5  & 5 & UN & 21.12\% & 51.10\% & 34.43\% \\ \midrule

\multicolumn{7}{l}{\textbf{Retriever (Vanilla \& Finetuned)}} \\
BGE        & --         & 5 & UF & 4.11\% & 13.14\% & 9.61\% \\
DPR        & --         & 5 & UN & 3.24\% & 9.38\%  & 7.29\% \\
FT-DPR     & --         & 5 & UN & 2.60\% & 8.22\%  & 6.31\% \\
ColBERT    & --         & 5 & UN & 3.82\% & 11.86\% & 8.28\% \\
FT-ColBERT & --         & 5 & UF & 3.36\% & 8.85\%  & 6.42\% \\ \midrule

\multicolumn{7}{l}{\textbf{Retriever (Vanilla \& Finetuned) + Reranker (Vanilla)}} \\
BGE        & MonoT5     & 5 & UN & 4.98\% & 15.34\% & 12.38\% \\
DPR        & MonoT5     & 5 & UN & 4.22\% & 11.75\% & 9.20\% \\
FT-DPR     & MonoT5     & 3 & UF & 4.46\% & 12.50\% & 9.90\% \\
ColBERT    & MonoT5     & 3 & UF & 4.46\% & 11.98\% & 8.46\% \\
FT-ColBERT & MonoT5     & 3 & UF & 4.22\% & 11.75\% & 8.45\% \\ \midrule

\multicolumn{7}{l}{\textbf{Retriever (Vanilla \& Finetuned) + Reranker (Finetuned)}} \\
BGE        & FT-MonoT5  & 5 & UF & 4.86\% & 13.89\% & 10.82\% \\
DPR        & FT-MonoT5  & 3 & UF & 4.10\% & 10.94\% & 8.75\% \\
FT-DPR     & FT-MonoT5  & 3 & UN & 4.17\% & 12.44\% & 8.80\% \\
ColBERT    & FT-MonoT5  & 5 & UF & 4.46\% & 11.81\% & 8.68\% \\
FT-ColBERT & FT-MonoT5  & 5 & UN & 4.22\% & 11.69\% & 7.87\% \\ \bottomrule

\end{tabular}%
}
\end{table}

We also evaluate rerankers under an oracle setting, where we assume the retriever is perfect and can retrieve all relevant passages. In this setup, we compare the vanilla and fine-tuned versions of MonoT5. Table~\ref{tbl:oracle_results} shows that fine-tuned MonoT5 achieves the best performance when retrievers are assumed to be oracle. This suggests that fine-tuning can help rerankers in principle, but the improvements are still small, even under ideal retrieval. Thus, the poor performance of fine-tuned MonoT5 in Table~\ref{tbl:retrievers_finetuned_reranker} may be attributed to the limitations of the retrievers rather than rerankers themselves. Nonetheless, the results confirm that fine-tuning alone is insufficient, and more advanced approaches are needed to handle answer-supporting passages effectively.

\subsection{Reader}\label{ss:reader}

After retrieving and reranking passages, we use the top-1, top-3, and top-5 passages as input to the reader in the RAG setting. This setup allows us to evaluate retrievers and rerankers on the Inferential QA task. A key challenge in the top-3 and top-5 settings is that simply concatenating passages to form the context introduces redundancy, since multiple passages may contain overlapping hints\footnote{This overlap occurs because, as discussed in Section~\ref{ss:dataset_preparation}, relevant passages for a question are generated from the subsets and permutations of the five hints designed for that question.}. To address this, we propose two methods for constructing the final context: (1) $\text{Union}_{norm}$ and (2) 
$\text{Union}_{freq}$. 

We represent each passage $p_i = \{s_{i1}, \dots, s_{ij}\}$, where $s_{ij}$ denotes the $j$-th sentence of passage $p_i$. The $\text{Union}_{norm}$ method applies a standard union while preserving order, producing a new set of sentences that are concatenated into a passage. The $\text{Union}_{freq}$ method instead scores each sentence based on two factors: (i) the rank of the passage it belongs to ($rank(P_i)$), and (ii) the position of the sentence within that passage ($pos_i(s)$). The scoring function is defined as:

\begin{equation}\label{eq:union_freq}
\text{score}(s) \;=\;
\alpha \cdot \sum_{i:\,s \in P_i} \frac{1}{\text{rank}(P_i)}
\;+\;
\beta \cdot \sum_{i:\,s \in P_i} \frac{1}{\text{pos}_i(s)}
\end{equation}

where $\alpha$ controls the weight given to higher-ranked passages and $\beta$ controls the weight given to earlier sentences within each passage. In our experiments, we set $\alpha = 0.6$ and $\beta = 0.4$\footnote{Grid search on the development set indicated that $\alpha=0.6$ and $\beta=0.4$ achieve the highest EM.}.
After computing the scores, sentences are ranked in descending score order, and the top-$k$ are concatenated to form a passage.

\begin{figure}[t]
  \centering
  \includegraphics[width=\columnwidth]{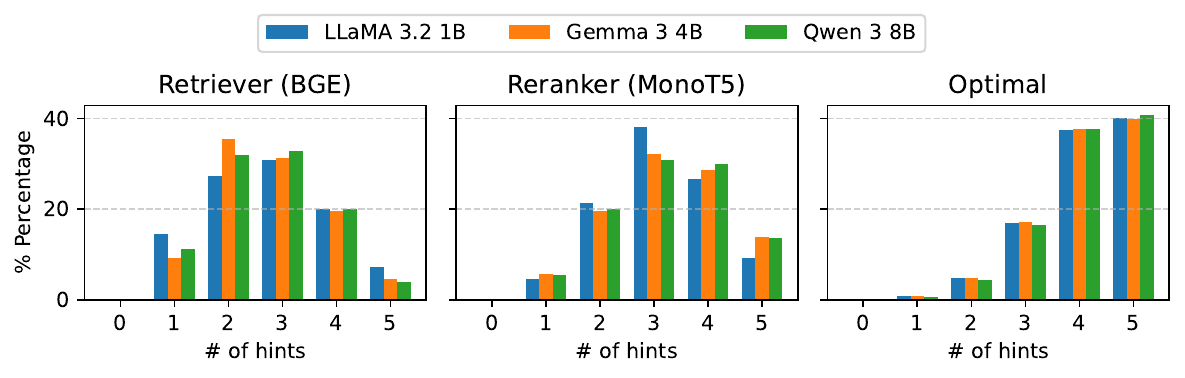}
    \caption{Distribution of passages by the number of hints across three settings: Retriever without Reranker, Oracle Retriever with Reranker, and Optimal.}

    \label{fig:hint_contribution}
\end{figure}

Table~\ref{tbl:rag_results} reports the results of different retriever and reranker setups in LLM-based Reader. 
The \textit{Optimal} setting corresponds to an oracle scenario where both retriever and reranker perform perfectly.
To simulate this setting, we exhaustively evaluate all fully and partially relevant passages for each question and mark a question as answerable if at least one of these passages enables the LLM to produce the correct answer, showing the upper bound achievable when using LLaMA 3.2 1B, Gemma 3 4B, and Qwen 3 8B as readers. These results demonstrate that if retrieval and reranking were perfectly solved, EM could reach as high as 90.16\% with Gemma 3 4B, and similarly high values for the other LLMs. This indicates that \dataset is of high quality and can effectively support LLMs in generating correct answers. 

However, under the \textit{Oracle retriever} setting—where we assume perfect retrieval but rely on rerankers to order passages—the EM is only about half of the \textit{Optimal} score, showing that current rerankers are not capable of ordering passages effectively. Fine-tuning also fails to close this gap, indicating that rerankers require new methods and paradigms tailored to the Inferential QA task. Moreover, retriever-only results are the weakest, far below the \textit{Optimal} scores, highlighting that current retrievers are also inadequate and must move beyond existing paradigms. While rerankers provide slight improvements, they cannot resolve the fundamental challenges. As shown in Figure~\ref{fig:rag_results}, fine-tuning retrievers can even degrade performance, while rerankers provide only marginal improvements. 

Table~\ref{tbl:retriever_results} shows that the fine-tuned DPR achieves better performance than DPR-vanilla; however, in the LLM-based Reader stage, its EM is lower than that of the vanilla model. A possible explanation is that fine-tuned retrievers may retrieve more relevant passages overall, but their ranking is suboptimal. In this case, rerankers can help improve their ordering, leading to better results than the vanilla model when rerankers are applied. 
Fine-tuned rerankers perform slightly worse than their vanilla counterparts in the LLM-based Reader, as shown in Table~\ref{tbl:retrievers_finetuned_reranker}, indicating that they provide little benefit. However, this figure shows that combining a fine-tuned retriever with a fine-tuned reranker yields better performance than other setups, although the improvement is very small. 

The results show that general-purpose LLMs, such as Gemma 3 4B, outperform reasoning-oriented LLMs like Qwen 3 8B, even with fewer parameters. This indicates that relying on reasoning-oriented LLMs is not an effective solution for the Inferential QA task. Overall, \textit{the results confirm that current retrievers and rerankers are not well-suited for the Inferential QA task, and that fine-tuning alone is not a viable solution; entirely new approaches are needed.}

Figure~\ref{fig:hint_contribution} illustrates this challenge. Retrievers and rerankers perform best when identifying passages with three sentences, whereas the \textit{Optimal} results show that the most helpful passages for LLM-based Reader contain five hints. This is because longer passages provide more information about the answer, enabling the LLM-based Reader to answer with higher confidence. The weaker performance of retrievers and rerankers on passages with four or five hints may explain, in part, why they struggle with passages.

\section{Conclusion}\label{s:conclusion}

In this paper, we introduced \textbf{Inferential QA}—a new reasoning QA task that challenges Question Answering (QA) systems to infer answers from \emph{answer-supporting} passages rather than extract them verbatim. To enable this study, we constructed \textbf{\dataset} (\textbf{QU}estions requiring \textbf{I}nference from \textbf{T}exts), a large-scale dataset of 7,401 questions and 2.4 million passages derived from high-convergence hints, labeled through a combination of LLM-based answerability and human verification.
Our extensive evaluation across multiple retrievers, rerankers, and LLM-based readers reveals that existing QA methods—effective on traditional answer-containing datasets—struggle under inferential conditions. Retrievers often fail to identify the correct answer-supporting passages, rerankers provide only minor improvements, and fine-tuning leads to limited or inconsistent gains. Even reasoning-oriented LLMs do not outperform smaller general-purpose models, suggesting that inference-based reasoning remains a major bottleneck in current QA pipelines.
These findings highlight a fundamental gap between extraction and inference in QA. By shifting focus toward questions that require reasoning over indirect clues, \textbf{Inferential QA} opens a new research direction for developing retrieval, reranking, and reasoning methods capable of drawing conclusions from subtle, distributed evidence. 
We believe this work paves the path towards future research on designing retrieval-augmented reasoning systems that more closely resemble human inferential comprehension.

\begin{acks}
The computational results presented here have been achieved using the LEO HPC infrastructure of the University of Innsbruck.
This work was supported in part by the Center for Intelligent Information Retrieval, in part by NSF grant \#2143434, and in part by the Office of Naval Research contract \#N000142412612. Any opinions, findings and conclusions or recommendations expressed in this material are those of the authors and do not necessarily reflect those of the sponsor.
\end{acks}

\bibliographystyle{ACM-Reference-Format}
\balance
\bibliography{Main}

\appendix

\section{Appendix}\label{apx}

\subsection{Human Verification}\label{apx:human_verification}
\begin{figure}[h]
  \centering
  \includegraphics[width=0.8\columnwidth]{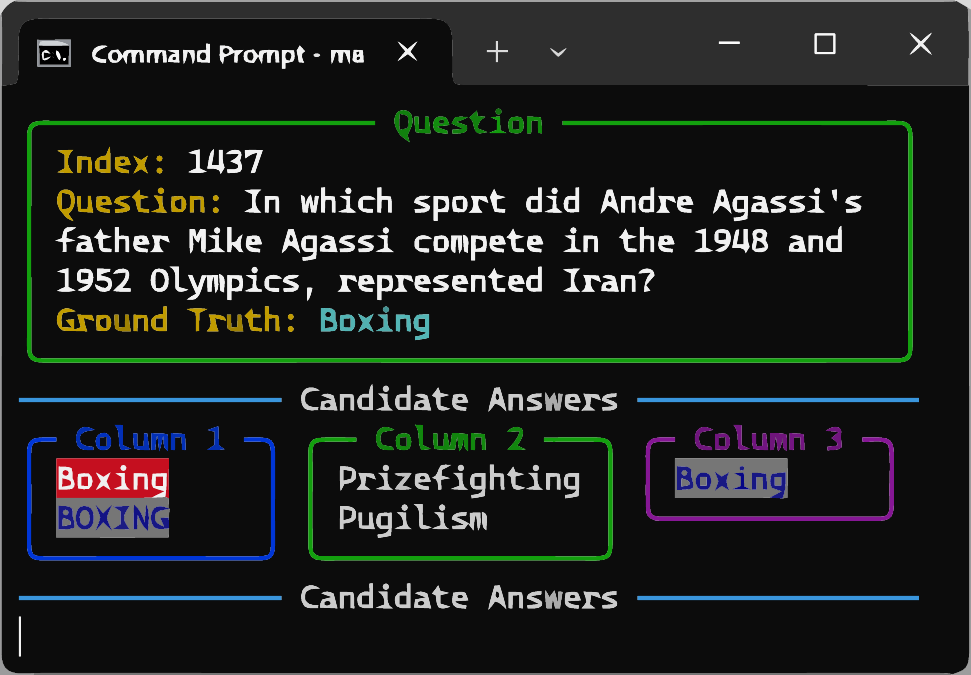}
    \caption{The human annotation interface. The \textit{Question} section displays the question and its correct answer, while the \textit{Candidate Answers} section lists all answers generated by LLMs across 325 passages. Annotators manually verify and select the correct answers.}
    \label{fig:human_evaluation}
\end{figure}
\begin{table}[h]
	\centering
	
	\caption{Demographic information of evaluators}
	\resizebox{\columnwidth}{!}{%
		\scriptsize
		\begin{tabular}{@{}c@{\hspace{35pt}}c@{\hspace{35pt}}c@{\hspace{35pt}}c@{}}
			\toprule
			\textbf{Evaluator} & \textbf{Gender} & \textbf{Age} & \textbf{Education Level} \\ \midrule
			1                  & Female          & 19           & High School              \\
			2                  & Female          & 30           & Bachelor's Degree             \\
			3                  & Male          & 40           & Master's Degree     \\
			4                  & Male            & 32           & PhD                      \\ \bottomrule
		\end{tabular}%
	}
	\label{tbl:evaluator_background}
\end{table}

To improve the reliability of the development and test sets, we perform human verification of answers marked as correct by GPT-Eval, mitigating potential errors or hallucinations from LLMs. Annotators use a custom interface shown in Figure~\ref{fig:human_evaluation} to validate answer correctness and resolve ambiguous cases.

Specifically, we display all generated answers that GPT-Eval marked as correct and highlight those that exactly match the ground-truth answers extracted from their original sources (TriviaHG~\cite{10.1145/3626772.3657855} and WikiHint~\cite{10.1145/3726302.3730284}). Human annotators are then asked to review these answers and select or deselect the ones they believe have been correctly or incorrectly labeled. For each question, annotators see the question text, the generated answers, and the associated ground-truth answer(s). They are instructed to assess whether the generated answers are semantically equivalent to the ground-truth answer, even if they differ lexically (e.g., \textit{“USA”} vs. \textit{“United States of America”}). If a generated answer is deemed incorrect despite being accepted by GPT-Eval, it is manually unmarked. Conversely, answers that are correct but not highlighted can be selected to ensure no valid answers are missed. This process helps correct both false positives and false negatives, resulting in higher-quality passage labels for downstream evaluation. Table~\ref{tbl:evaluator_background} summarizes the demographic background of the human annotators involved in this verification process.

\subsection{Finetuned Retrievers}\label{apx:finetuned_retrievers}
To fine-tune ColBERT and DPR, we experiment with different numbers of positive and negative samples. Given the complexity and reasoning requirements of answer-supporting passages, we hypothesize that even a single positive example might be sufficient for effective fine-tuning. To explore this, we test a range of values—$1$, $5$, $10$, $50$, $100$, and $200$—for both positive and negative samples across ColBERT and DPR, aiming to identify the most optimal configuration.

The results are presented in Table~\ref{tbl:finetuned_colbert} and Table~\ref{tbl:finetuned_dpr}, which report performance for each combination. Based on these results, we find that using 50 positive samples yields the best performance for ColBERT, while 10 positive samples perform best for DPR. We therefore adopt these configurations as the final fine-tuned versions of ColBERT and DPR used in our experiments.

\begin{table}[b!]
\small
\centering
\caption{Performance of finetuned ColBERT retrievers with varying numbers of positive passages. We report Hit@$k$, Recall@$k$, MRR, and nDCG@$k$. The best are highlighted.}
\label{tbl:finetuned_colbert}
\resizebox{\columnwidth}{!}{%
\scriptsize
\begin{tabular}{@{}l c c c c c c c c}
\toprule
\textbf{Retriever}
& \multicolumn{2}{c}{Hit@K}
& \multicolumn{3}{c}{Recall@K}
& MRR@K
& \multicolumn{2}{c}{nDCG@K} \\
\cmidrule(lr){2-3} \cmidrule(lr){4-6} \cmidrule(lr){7-7} \cmidrule(lr){8-9}
& 1 & 5
& 5 & 10 & 50
& 100
& 10 & 100 \\ \midrule

\multicolumn{9}{l}{\textbf{Vanilla}} \\
ColBERT        & 20.31\% & 21.12\% & 0.31\% & 0.63\% & 3.15\% & 20.69\% & 16.14\% & 17.51\% \\ \midrule

\multicolumn{9}{l}{\textbf{Finetuned}} \\
ColBERT (1 pos)   & 13.95\% & 17.88\% & 0.21\% & 0.42\% & 1.95\% & 15.93\% & 10.40\% & 10.36\% \\
ColBERT (5 pos)   & 15.74\% & 20.14\% & 0.24\% & 0.46\% & 2.20\% & 17.93\% & 11.61\% & 11.93\% \\
ColBERT (10 pos)  & 16.90\% & 20.37\% & 0.25\% & 0.49\% & 2.40\% & 18.64\% & 12.21\% & 13.20\% \\
ColBERT (50 pos)  & \highlight{21.11\%} & \highlight{23.84\%} & \highlight{0.27\%} & \highlight{0.53\%} & \highlight{2.56\%} & \highlight{20.83\%} & \highlight{13.00\%} & \highlight{13.98\%} \\
ColBERT (100 pos) & 18.46\% & 22.92\% & 0.29\% & 0.56\% & 2.75\% & 20.56\% & 13.80\% & 14.87\% \\
ColBERT (200 pos) & 17.77\% & 23.32\% & 0.28\% & 0.55\% & 2.69\% & 20.35\% & 13.21\% & 14.58\% \\ \bottomrule
\end{tabular}%
}
\end{table}

\begin{table}[b!]
\small
\centering
\caption{Performance of finetuned DPR retrievers with varying numbers of positive passages. We report Hit@$k$, Recall@$k$, MRR, and nDCG@$k$. The best are highlighted.}
\label{tbl:finetuned_dpr}
\resizebox{\columnwidth}{!}{%
\scriptsize
\begin{tabular}{@{}l c c c c c c c c}
\toprule
\textbf{Retriever}
& \multicolumn{2}{c}{Hit@K}
& \multicolumn{3}{c}{Recall@K}
& MRR@K
& \multicolumn{2}{c}{nDCG@K} \\
\cmidrule(lr){2-3} \cmidrule(lr){4-6} \cmidrule(lr){7-7} \cmidrule(lr){8-9}
& 1 & 5
& 5 & 10 & 50
& 100
& 10 & 100 \\ \midrule

\multicolumn{9}{l}{\textbf{Vanilla}} \\
DPR        & 18.58\% & 21.18\% & 0.28\% & 0.54\% & 2.54\% & 19.77\% & 13.68\% & 13.64\% \\ \midrule

\multicolumn{9}{l}{\textbf{Finetuned}} \\
DPR (1 pos)   & 19.97\% & 24.83\% & 0.31\% & 0.61\% & 2.97\% & 22.42\% & 14.69\% & 16.05\% \\
DPR (5 pos)   & 20.83\% & 25.98\% & 0.31\% & 0.62\% & 3.08\% & 23.50\% & 14.61\% & 16.44\% \\
DPR (10 pos)  & \highlight{20.91\%} & \highlight{28.07\%} & \highlight{0.32\%} & \highlight{0.63\%} & \highlight{3.12\%} & \highlight{23.56\%} & \highlight{14.98\%} & \highlight{16.69\%} \\
DPR (50 pos)  & 20.31\% & 26.97\% & 0.31\% & 0.62\% & 3.05\% & 23.70\% & 14.49\% & 16.17\% \\
DPR (100 pos) & 20.83\% & 27.43\% & 0.31\% & 0.61\% & 3.03\% & 24.19\% & 14.14\% & 15.86\% \\
DPR (200 pos) & 20.06\% & 25.75\% & 0.30\% & 0.60\% & 2.93\% & 24.07\% & 13.96\% & 15.36\% \\ \bottomrule
\end{tabular}%
}
\end{table}

\subsection{Finetuned Reranker}\label{apx:finetuned_reranker}

\begin{table}[b]
\small
\centering
\caption{Performance of finetuned MonoT5 rerankers with varying numbers of positive passages. We report nDCG@$k$.}
\label{tbl:finetuned_monot5}
\resizebox{\columnwidth}{!}{%
\scriptsize
\begin{tabular}{@{} p{70pt} l c c c c c c@{}}
\toprule
\textbf{Reranker}
& \multicolumn{6}{c}{nDCG@K} \\
\cmidrule(lr){2-7}
& 1 & 5 & 10 & 20 & 50 & 100 \\ \midrule

\multicolumn{7}{l}{\textbf{Vanilla}} \\
MonoT5        & 80.83\% & 82.01\% & 82.95\% & 84.24\% & 86.46\% & 88.71\% \\ \midrule

\multicolumn{7}{l}{\textbf{Finetuned}} \\
MonoT5 (1 pos)   & 71.72\% & 76.07\% & 78.35\% & 80.61\% & 83.88\% & 86.78\% \\
MonoT5 (5 pos)   & 81.91\% & 82.94\% & 83.61\% & 84.67\% & 86.65\% & 88.87\% \\
MonoT5 (10 pos)  & \highlight{82.95\%} & \highlight{83.56\%} & \highlight{84.24\%} & \highlight{85.30\%} & \highlight{87.08\%} & \highlight{89.17\%} \\
MonoT5 (50 pos)  & 75.42\% & 77.87\% & 79.49\% & 81.29\% & 84.09\% & 86.92\% \\
MonoT5 (100 pos) & 75.58\% & 77.45\% & 78.73\% & 80.47\% & 83.47\% & 86.43\% \\ \bottomrule
MonoT5 (200 pos) & 75.12\% & 77.14\% & 78.08\% & 80.07\% & 82.71\% & 85.44\% \\ \bottomrule
\end{tabular}%
}
\end{table}

To fine-tune rerankers, we experiment with MonoT5, as it is a widely used reranking model and serves as a strong baseline. Similar to retrievers, we test different numbers of positive and negative samples—$1$, $5$, $10$, $50$, $100$, and $200$—to analyze the effect of training data size on reranking answer-supporting passages.

The results are presented in Table~\ref{tbl:finetuned_monot5}, which reports performance for each configuration. Based on these results, we find that using 10 positive samples provides the best performance for MonoT5. We therefore adopt this configuration as the final fine-tuned version of MonoT5 used in our experiments.

\subsection{Few-shot Prompt}\label{apx:prompt}

\begin{figure*}[t]
\centering
\begin{minipage}{0.96\textwidth}
\begin{tcolorbox}[
  colback=gray!5,
  colframe=gray!40,
  boxrule=0.4pt,
  arc=2pt,
  left=6pt,
  right=6pt,
  top=6pt,
  bottom=6pt,
  fonttitle=\bfseries,
  enhanced,
]
\small
\textbf{System:} You are an assistant that answers questions based on the provided context. You just answer questions with exact answers. You do not use sentences as the response.

\vspace{0.5em}
\textbf{User:}
Use the context to answer the question under conditions:
1. Answer should not be sentences. It should be some words. 
2. Do not generate "sorry" or "I cannot ..." sentences; instead, use "NO ANSWER". 
3. Do not generate explanations, reasoning, or full sentences—only provide the exact answer. 
4. If the answer cannot be guessed from the context, respond only with "NO ANSWER".

\vspace{0.8em}
\textbf{Context:}
He was the 44th President of the United States.  
He served as President from 2009 to 2017.  
He was the first African-American President of the United States.  
He was a member of the Democratic Party.  
He was born on August 4, 1961 in Honolulu, Hawaii.

\textbf{Question:} Who won the Nobel Peace Prize in 2009? 

\vspace{0.5em}
\textbf{Assistant:} Barack Obama

\vspace{0.8em}
\textbf{Context:}
The capital city of this country is Paris.  
This country is located in northwestern Europe.  
This country has a long history and has played a significant role in international affairs.  
The official language of this country is French.  
The currency used in this country is the Euro.

\textbf{Question:} Édouard Daladier became Prime Minister of which country in 1933?  

\vspace{0.5em}
\textbf{Assistant:} France

\vspace{0.8em}
\textbf{Context:}
It's the coldest season of the year.  
It's the season when snow falls in many regions.  
It's the season when many people celebrate Christmas and New Year's Eve.  
It's the season when days are shorter and nights are longer.  
It's the season when many animals hibernate.

\textbf{Question:} If you have a 'Mahonia Japonica', in which season will it be in flower?  

\vspace{0.5em}
\textbf{Assistant:} Winter

\vspace{0.8em}
\textbf{Context:}
It is a team sport that originated in the United States.  
It is played with an oval-shaped ball.  
The objective of the game is to score points by advancing the ball into the opposing team's end zone.  
Points can be scored by carrying the ball across the opponent's goal line, throwing it to a teammate in the end zone, or kicking it through the opponent's goalposts.  
The game is divided into four quarters, each lasting 15 minutes.

\textbf{Question:} Which sport is played under the 'Harvard Rules'?  

\vspace{0.5em}
\textbf{Assistant:} AMERICAN FOOTBALL

\vspace{0.8em}
\textbf{Context:}
He was born on April 20, 1889, in Braunau am Inn, Austria.  
He was the leader of the Nazi Party.  
He became the chancellor of Germany in 1933.  
He took the title of Führer und Reichskanzler in 1934.  
He initiated World War II in Europe by invading Poland on September 1, 1939.

\textbf{Question:} Who was made an honorary citizen of Haslach, Austria, in 1938, an honour withdrawn in 2004?  

\vspace{0.5em}
\textbf{Assistant:} Adolf Hitler

\vspace{0.8em}
\textbf{Context:}
Its capital is Beijing.  
Its population is more than 1 billion.  
This country has a history spanning more than 3,000 years of continuous civilization.

\textbf{Question:} Which country borders 14 others and uses a single time zone across its vast territory?  

\vspace{0.5em}
\textbf{Assistant:} \textcolor{red}{\textit{China}}
\end{tcolorbox}
\end{minipage}
\caption{Few-shot prompt provided to the language model during inference. Each example demonstrates how to use hint-based context to answer questions concisely with short responses (or \texttt{NO ANSWER} if the context is insufficient). The final pair shows the new query to be answered, with the model-generated answer shown in \textcolor{red}{red} for clarity.}
\label{fig:full_prompt}
\end{figure*}

Figure~\ref{fig:full_prompt} illustrates the few-shot prompt used to evaluate language models. 
It comprises several \textit{question–context–answer} examples, where each context is constructed by concatenating hints that indirectly describe an entity, concept, or event. 
The prompt demonstrates how to infer answers from contextual clues without generating explanations, guided by a system instruction that enforces short, phrase-level outputs, the use of ``\texttt{NO ANSWER}'' when information is insufficient, and the avoidance of reasoning or justification. 
The final example presents a new inferential question with unseen hints, prompting the model to produce its answer following the learned pattern.

\end{document}